%% file: main.tex
\documentclass[runningheads]{llncs}
\usepackage{graphicx}
\usepackage{tikz}
\usepackage{comment}
\usepackage{amsmath,amssymb} % define this before the line numbering.
\usepackage{color}

\usepackage{multirow}
\usepackage{algorithm}
\usepackage{algorithmic}
\usepackage{subcaption}
\usepackage{booktabs}
\usepackage{xspace}

\renewcommand{\algorithmiccomment}[1]{\bgroup\hfill//~#1\egroup}
\newcommand{\STAB}[1]{\begin{tabular}{@{}c@{}}#1\end{tabular}}

\def\eg{\emph{e.g}.\xspace}

\def\etal{\emph{et al}.\xspace}

\usepackage[accsupp]{axessibility}  % Improves PDF readability for those with disabilities.

\begin{document}
% \linenumbers
\pagestyle{headings}
\mainmatter
\def\ECCVSubNumber{4329}  % Insert your submission number here

\title{Improving Vision Transformers for \\ Incremental Learning} % Replace with your title

\begin{comment}
\titlerunning{ECCV-22 submission ID \ECCVSubNumber} 
\authorrunning{ECCV-22 submission ID \ECCVSubNumber} 
\author{Anonymous ECCV submission}
\institute{Paper ID \ECCVSubNumber}
\end{comment}
%******************

%\begin{comment}
\titlerunning{Improving Vision Transformers for Incremental Learning}
\author{Pei Yu \and
Yinpeng Chen \and
Ying Jin \and
Zicheng Liu}
\authorrunning{P. Yu et al.}
\institute{Microsoft\\
\email{\{peyu, yiche, yinjin, zliu\}@microsoft.com}}
%\institute{Microsoft}
%\end{comment}
%******************
\maketitle

\begin{abstract}
This paper proposes a working recipe of using Vision Transformer (ViT) in class incremental learning. Although this recipe only combines existing techniques, developing the combination is not trivial. Firstly, naive application of ViT to replace convolutional neural networks (CNNs) in incremental learning results in serious performance degradation. Secondly, we nail down three issues of naively using ViT: (a) ViT has very slow convergence when the number of classes is small, (b) more bias towards new classes is observed in ViT than CNN-based architectures, and (c) the conventional learning rate of ViT is too low to learn a good classifier layer. Finally, our solution, named ViTIL (ViT for Incremental Learning) achieves new state-of-the-art on both CIFAR and ImageNet datasets for all three class incremental learning setups by a clear margin. We believe this advances the knowledge of transformer in the incremental learning community. Code will be publicly released.
%\dots
\end{abstract}

%%%%%%%%% BODY TEXT
\input{introduction}
\input{relatedwork}
\input{technical}
\input{experiment}
\input{conclusion}

\clearpage
\bibliographystyle{splncs04}
\bibliography{egbib}

\appendix
\input{appendix}

\end{document}

%% file: introduction.tex
\section{Introduction}
\label{sec:Intro}

\begin{figure}[t]
	\centering
	\includegraphics[width=\linewidth]{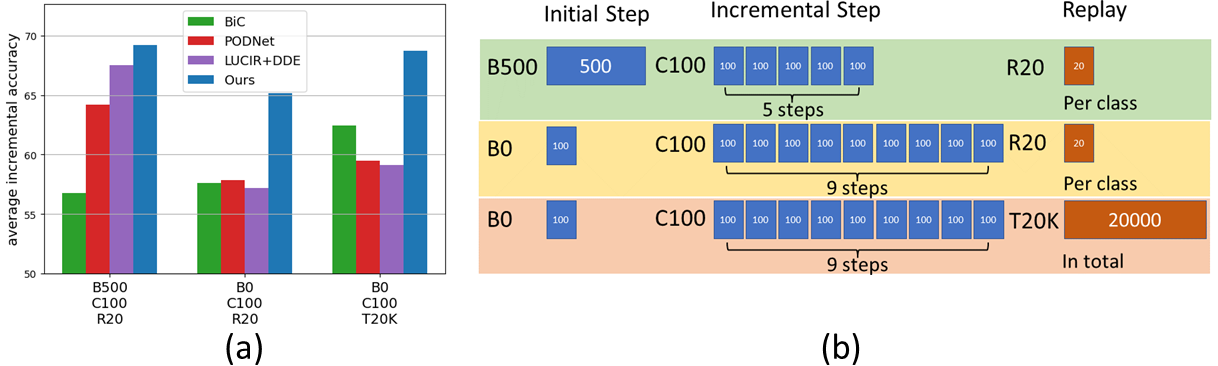}
	\caption{{\bf Comparison with state-of-the-art methods on three CIL setups} on ImageNet-1000. (a) shows average incremental accuracy. (b) illustrates CIL setups. B500 denotes initial step contains half of total 1000 classes. B0 denotes model starts from scratch, and each step adds same number of new classes. C100 denotes each incremental step adds 100 new classes. R20 denotes each old class keeps same number of 20 exemplars. T20K denotes each incremental step keeps total 20000 exemplars for all classes. (Best viewed in color)}
	\label{fig:Teaser}
\end{figure}

\begin{figure*}[t]
	\centering
	\begin{subfigure}{0.328\linewidth}
		\includegraphics[width=\linewidth]{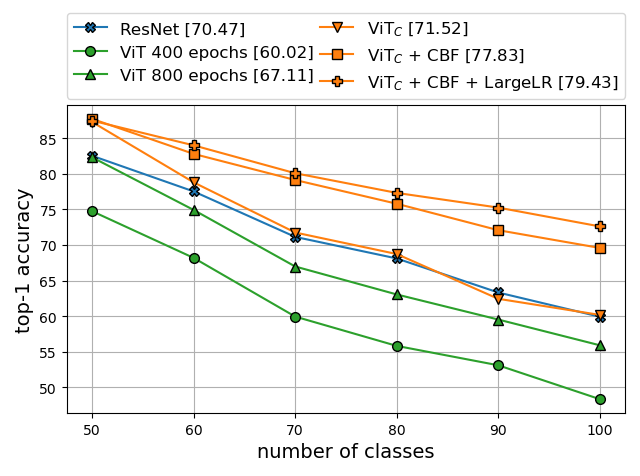}
		\caption{B50, C10, R20}
		\label{fig:NaiveViTOurs:fg50p20}
	\end{subfigure}
	\hfill
	\begin{subfigure}{0.328\linewidth}
		\includegraphics[width=\linewidth]{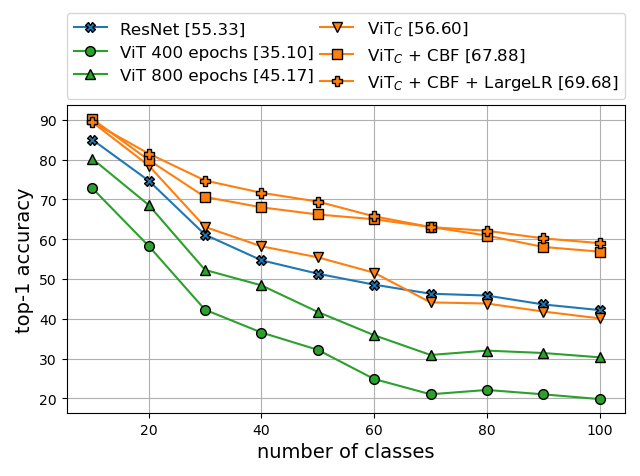}
		\caption{B0, C10, R20}
		\label{fig:NaiveViTOurs:fg10p20}
	\end{subfigure}
	\hfill
	\begin{subfigure}{0.328\linewidth}
		\includegraphics[width=\linewidth]{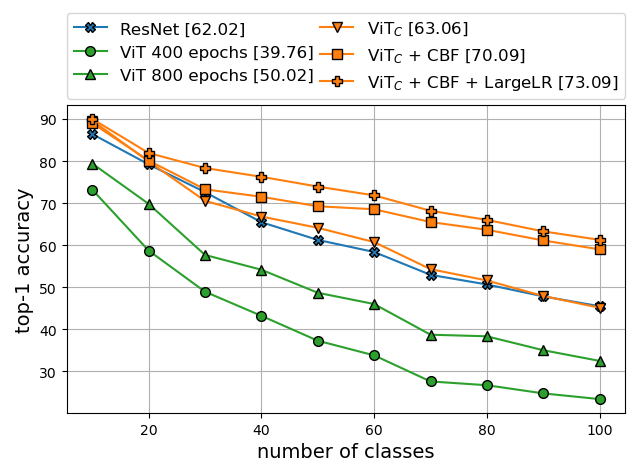}
		\caption{B0, C10, T2K}
		\label{fig:NaiveViTOurs:fg10p200}
	\end{subfigure}
	\caption{{\bf Comparison among LUCIR, naive ViT, and our method on incremental accuracy of ImageNet-100, with three different CIL protocols}. B50 denotes initial step contains 50 classes. C10 denotes each incremental step adds 10 new classes. R20 denotes each old class keeps 20 exemplars. ``ResNet" denotes LUCIR with ResNet-18. ``ViT" represents replacing ResNet-18 in LUCIR naively with ViT-ti model, trained for 400 or 800 epochs. ``ViT$_{C}$ + CBF + LargeLR" denotes our method using ViT-ti model with convolutional stem (ViT$_{C}$), with bias correction by class balance finetuning (CBF), and large learning rate for classifier (LargeLR). Number in $[\ ]$ denotes average incremental accuracy. Results show that our method consistently outperforms ResNet-18 method LUCIR, and naive ViT method, by a large margin. Moreover, each component of our method provides improvement consistently. (Best viewed in color)}
	\label{fig:NaiveDeiT}
\end{figure*}

Recent progress in Vision Transformer \cite{dosovitskiy2021image} demonstrates superior performance over convolutional neural networks (CNNs) on various computer vision tasks such as image recognition \cite{touvron2021training,liu2021swin} and object detection~\cite{carion2020end}. The success of ViT motivates us to investigate if it is also suitable for class incremental learning (CIL), which is another great interest to the community since its setting is close to real-world situation, where the model needs to handle sequentially incoming data of new classes and avoid performance degradation on previous classes. The key question to answer is whether ViT provides a better feature extractor than CNNs for class incremental learning?

However, applying ViT in class incremental learning is not trivial. Naively replacing CNN feature extractor with ViT results in significant performance degradation across different CIL settings, as shown in Fig.~\ref{fig:NaiveDeiT}. After careful analysis, we found three factors contribute to the degradation: (a) ViT models have slow convergence, especially at the beginning of incremental learning where the number of classes is small, (b) more bias towards new classes is observed in ViT models than CNN models, mainly because the margin ranking loss (an effective bias removal technique for CNN) conflicts with data augmentation such as MixUp or CutMix which are important for ViT, and (c) using the same learning rate for both ViT feature extractor and classifier causes underfitting in classifier, which is reflected in the magnitude of learnable temperature in softmax.

Base upon our analysis, we further address these issues by simply using existing techniques in either network architecture or incremental learning. Firstly, we address the slow convergence by using convolutional stem \cite{xiao2021early} to replace the patchify stem in ViT model, which achieves better performance with significantly shorter training. Secondly, we find that finetuning with balanced dataset \cite{wu2019large,hou2019learning} is effective to correct bias towards new classes for vision transformer, but not conflicting with data augmentation (MixUp, CutMix, etc). Finally, we show that using larger learning rate for classifier results in larger value of softmax temperature, further boosting the performance. As shown in Fig.~\ref{fig:NaiveDeiT}, these techniques effectively boost performance across three CIL settings. Note that none of these techniques is originally proposed in this paper, our major contribution is to locate key issues in applying vision transformer in incremental learning and connects existing techniques to address these issues effectively.

With these techniques, as shown in Fig.~\ref{fig:Teaser} (a), our method ViTIL(ViT for Incremental Learning) achieves new state-of-the-art consistently for all three class incremental learning setups by a clear margin.
In contrast, existing methods only perform well in one CIL setup.
For instance, on ImageNet-1000, our ViTIL achieves 69.20\% top-1 accuracy for the CIL setup of 500 initial classes, and each incremental step adds 100 new classes, outperforming LUCIR+DDE~\cite{hu2021distilling} by $1.69\%$.
For CIL setup of 10 incremental steps (each step adds 100 new classes), our method outperforms PODNet~\cite{douillard2020podnet} by $7.27\%$ ($65.13\%$ vs $57.86\%$), when each old class keeps 20 exemplars.
When each incremental step keeps total 20000 exemplars, our method outperforms BiC~\cite{wu2019large} by $6.28\%$ ($68.75\%$ vs $62.47\%$).

In summary, our contributions are three-fold.
\begin{enumerate}
\item We nail down three key issues (slow convergence, bias towards new classes, underfitted classifier) that contributes to the degradation of applying vision transformer in class incremental learning.
\item We showcase a simple solution with existing techniques that effectively addresses all three issues, achieving significant performance boost.
\item Our method (ViTIL) achieves the new state-of-the-art across three class incremental setups by a clear margin. This is challenging as these setups previously have different lead methods.
\end{enumerate}

The rest of this paper is organized as follows.
Sec.~\ref{sec:RelatedWork} introduces related work.
In Sec.~\ref{sec:Baseline}, we investigate applying ViT naively to CIL, and its performance issue.
In Sec.~\ref{sec:OurMethod}, our method is proposed to address these issues.
Sec.~\ref{sec:Exp} evaluates the proposed method in different CIL setups, and studies the impact of different components of the proposed method.
Sec.~\ref{sec:Conclusion} concludes this paper.

%% file: relatedwork.tex
\section{Related Work}
\label{sec:RelatedWork}
Incremental learning methods can be categorized by how they tackle catastrophic forgetting.
Regularization-based methods penalize the changes of model parameters~\cite{kirkpatrick2017overcoming,zenke2017continual,lopez2017gradient,yoon2017lifelong,aljundi2018memory,chaudhry2018riemannian,ahn2019uncertainty,li2019learn,chaudhry2019efficient}.
Distillation-based methods tackle catastrophic forgetting by distilling knowledge from previous models~\cite{li2016learning,rebuffi2017icarl,aljundi2017expert,rannen2017encoder,hou2018lifelong,castro2018end,hou2019learning,wu2019large,dhar2019learning,lee2019overcoming,douillard2020podnet,liu2020mnemonics,hu2021distilling,zhao2020maintaining}.
Some methods finetune model with data exemplars of previous classes, without weight constraints or distillation~\cite{belouadah2019il2m,belouadah2020scail}.
Some methods seek synthetic data~\cite{shin2017continual,kemker2018fearnet}.
A more comprehensive survey can be found in~\cite{parisi2019continual}.

\vspace{1mm}
\noindent{\bf Distillation-based methods:}
Hinton \etal propose knowledge distillation to improve the performance of single model by distilling the knowledge from an ensemble of models~\cite{hinton2014distilling}.
In Learning without Forgetting (LwF)~\cite{li2016learning}, knowledge distillation is applied to address catastrophic forgetting under multi-task incremental learning scenario.
Following LwF, Aljundi \etal propose gating autoencoders to automatically select the task classifier~\cite{aljundi2017expert}.
In~\cite{rannen2017encoder}, distillation loss is added on the features from autoencoder.
In~\cite{hou2018lifelong}, knowledge is distilled from an expert CNN trained dedicated to new task.
In~\cite{dhar2019learning}, attention map is distilled.

\vspace{1mm}
\noindent{\bf CIL with data replay:}
In iCaRL~\cite{rebuffi2017icarl}, Rebuffi \etal show that, in addition to knowledge distillation, store a set exemplars from previous classes, and replay them in the following training process can significantly alleviate forgetting issue.
Nevertheless, bias towards new classes is a major issue in CIL with data replay.
Castro \etal propose to use balanced finetuning to tackle it~\cite{castro2018end}.
In~\cite{hou2019learning}, Hou \etal use normalized feature, normalized classifier, and margin ranking loss.
In~\cite{wu2019large}, bias correction is done by learning a bias correction layer.
In~\cite{zhao2020maintaining}, without an extra validation set, bias is mitigated by aligning the weight norm of new classes to old classes.
In~\cite{douillard2020podnet}, more sophisticated distillation is proposed by penalizing changes of spatially pooled features.
Instead of using herding to select exemplars, Liu \etal propose an automatic exemplar extraction framework mnemonics~\cite{liu2020mnemonics}.
~\cite{hu2021distilling} alleviates dependency on exemplars by distilling the causal data effect.

\vspace{1mm}
\noindent{\bf Transformers:}
Transformers were originally proposed for NLP tasks~\cite{vaswani2017attention}, and show superior performance for language model pre-training~\cite{radford2018improving,devlin2018bert}.
Increasing number of works seek to apply transformer to vision tasks, \eg, image recognition~\cite{touvron2021training,liu2021swin}, object detection~\cite{carion2020end}.
Vision transformer was proposed in~\cite{dosovitskiy2021image}, where input image is divided into non-overlapping patches simply with patchify operation.
Although it achieves superior performance, it replies on large corps of pre-training data.
In~\cite{touvron2021training}, Touvron \etal propose a data-efficient method for training Vision Transformer.
In~\cite{xiao2021early}, Xiao \etal seek to improve optimizability of Vision Transformer by using convolutional stem.

%% file: technical.tex
\section{Degradation of Applying ViT in CIL}
\label{sec:Baseline}
In this section, we show that a severe degradation is observed when naively appling Vision Transformers (ViT) in multi-class class incremental learning (CIL) with exemplars, and analyze the key degradation-causing factors.

\subsection{Review of Class Incremental Learning}
Class incremental learning (CIL)~\cite{rebuffi2017icarl,hou2019learning,wu2019large,douillard2020podnet,hu2021distilling} is performed progressively, consisting of incremental learning steps in which only the data of new classes is available. Particularly, at each incremental step $t$, a new model $\mathcal{M}_t$ is learned from an old model $\mathcal{M}_{t-1}$ and data of new classes, to perform classification on all classes it has seen.
Let us denote the old classes as $\mathcal{C}_{o}$ and the corresponding data as $\mathcal{X}_{o}=\{(x_i, y_i)| y_i\in\mathcal{C}_{o}\}$. Similarly, the data of new classes $\mathcal{C}_{n}$ is denoted as $\mathcal{X}_{n}=\{(x_i, y_i)| y_i\in\mathcal{C}_{n}\}$. We follow the setup of~\cite{rebuffi2017icarl,hou2019learning,wu2019large} to use the replay of old class exemplars, denoted by $\mathcal{X}^\prime_o\subset\mathcal{X}_{o}$.

In this paper, we follow LUCIR~\cite{hou2019learning} by replacing its CNN feature extractor with ViT model, but keeping its classifier that applies softmax on the cosine similarity as:
\begin{equation}
\label{equ:CosineSoftmax}
p_i(x)=\frac{\exp(\eta \left\langle\bar{\theta}_i, \bar{f}(x) \right\rangle)}{\sum_{j}\exp(\eta\left\langle\bar{\theta}_j, \bar{f}(x) \right\rangle)},
\end{equation}
where $\bar{\theta}, \bar{f}(x)$ denote $l_2$-normalized weight vector of each class, and the feature vector of input $x$, respectively. $\eta$ is a learnable softmax temperature parameter.

\subsection{Degradation of Replacing CNN with ViT}
\label{sec:Baseline:NaiveBase}
As shown in Fig.~\ref{fig:NaiveDeiT}, a severe performance degradation is observed when using ViT-tiny (ViT-ti) ~\cite{dosovitskiy2021image} to replace ResNet-18 as feature extractor, where the two models have similar computational cost (1.1G vs. 1.8G FLOPs). The average incremental accuracy drops by 10.45\%. The experiments are conducted on ImageNet-100, in which 50 classes are used to learn the first model and 5 incremental steps (10 classes for each) are followed. Interestingly, ViT model is behind ResNet-18 from the beginning (the first 50 classes) even though more training epochs are consumed (400 vs. 90 epochs). This seems not consistent with recent findings that ViT models outperform ResNet with similar FLOPs on ImageNet dataset. Next, we will discuss the causing factors.

\subsection{Analysis of Causing Factors}
\label{sec:Baseline:Diagnosis}
Our analysis locates three key issues related to the performance degradation of naive application of ViT, which will be discussed in this section.

\vspace{2mm}
\noindent{\bf Slow convergence when the number of classes is small:}
As ViT is significantly behind ResNet-18 for the first 50 classes, a reasonable guess is that ViT model is not fully trained even though it consumes longer training. 
As shown in Fig.~\ref{fig:DeiTConvstem}, ViT-ti takes more than 2000 training epochs to achieve saturated performance. 
In contrast, ResNet usually is trained for only 90 epochs.
When we increase training length from 400 to 800 epochs, it results in significant performance boosts from $74.72\%$ to $82.28\%$ (shown in Fig.~\ref{fig:DeiTConvstem}).
However, it is still lower than accuracy of ResNet-18. 
Although previous works~\cite{dosovitskiy2021image,touvron2021training} report that ViT models require longer training schedule (300-400 epochs) to achieve saturated performance, this issue becomes even more severe when the number of classes is small, which is typical in the beginning of incremental leaning.
As shown in Fig.~\ref{fig:DeiTConvstem}, ViT model requires more training epochs to achieve saturated performance on 10 classes than on 50 classes.
In addition, the initial model is crucial for the performance of the following incremental steps. 
As shown in Fig.~\ref{fig:NaiveViTOurs:fg50p20}, when the initial model is improved by training for 800 epochs, the average accuracy of all incremental steps is also significantly improved from $60.02\%$ to $67.11\%$.

\begin{figure*}[t]
	\centering
	\begin{subfigure}{0.328\linewidth}
		\includegraphics[width=\linewidth]{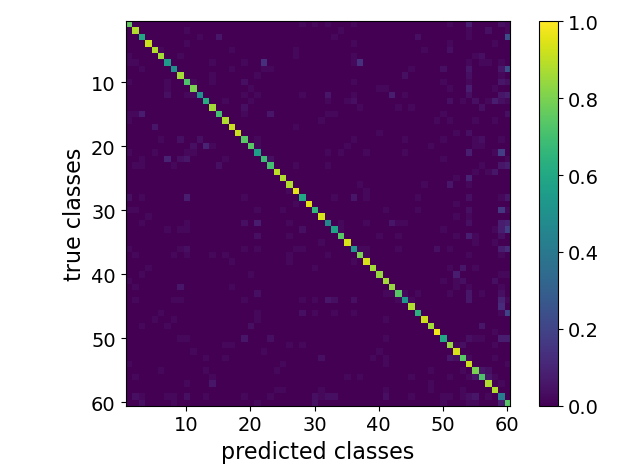}
		\caption{ResNet-18}
		\label{fig:ConfusionMatrixLUCIR}
	\end{subfigure}
	\hfill
	\begin{subfigure}{0.328\linewidth}
		\includegraphics[width=\linewidth]{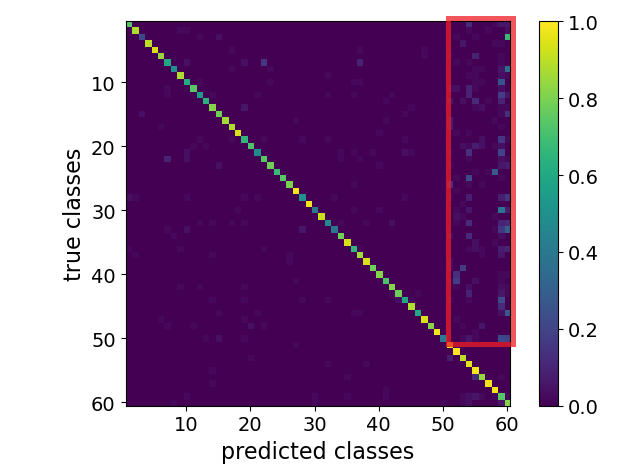}
		\caption{ViT-ti w/o CBF}
		\label{fig:ConfusionMatrixDeiT}
	\end{subfigure}
	\hfill
	\begin{subfigure}{0.328\linewidth}
		\includegraphics[width=\linewidth]{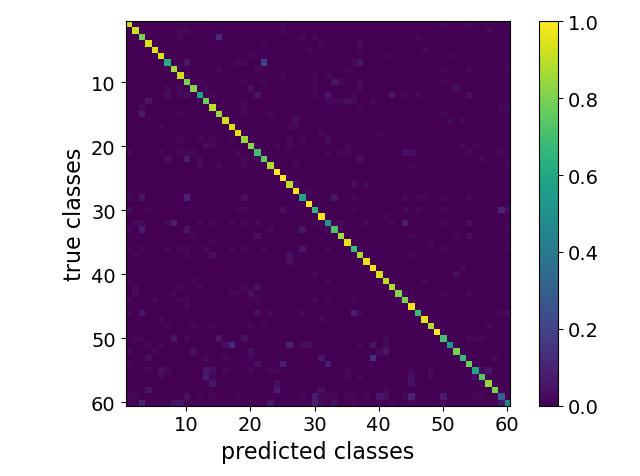}
		\caption{ViT-ti w/ CBF}
		\label{fig:ConfusionMatrixDeiTCBF}
	\end{subfigure}
	\caption{\textbf{Comparison between ResNet and ViT on confusion matrices} evaluated on ImageNet-100 after the first incremental step. Class 1-50 are old classes. Class 51-60 are new classes. \textbf{(a)} ResNet-18 with margin ranking loss, \textbf{(b)} ViT-ti model \textit{without} class balance finetuning (CBF), and \textbf{(c)} ViT-ti model \textit{with} class balance finetuning. (Best viewed in color)}
	\label{fig:ConfusionMatrices}
\end{figure*}

\vspace{2mm}
\noindent{\bf More bias towards new classes:}
We also compare ViT and ResNet on confusion matrices in Fig.~\ref{fig:ConfusionMatrices}, where clearly more bias towards new classes is observed in ViT (highlighted in a \textcolor{red}{red} box). Quantitatively, compared to ResNet-18, ViT-ti has 9.3\% more testing samples from old classes are falsely classified as new classes (11.2\% vs. 20.5\%).
We believe this is caused by the conflict between the margin ranking loss and the data augmentation like Mixup or CutMix. Margin ranking loss \cite{hou2019learning} is an important technique to prevent bias, but requires hard class label per sample. Such hard class label is not available in ViT as it heavily relies on augmentation like Mixup or CutMix that generates soft (or mixed) label. 
In contrast, ResNet is not affected as strong augmentation is not necessary.
Note that to avoid bias caused by long training schedule of ViT, we have already reduced training length for incremental steps $t>1$.
More details about training setting are described in Sec.~\ref{sec:ImpleDetails}.

\begin{figure}[t]
	\begin{minipage}[c]{0.47\linewidth}
	\centering
	\includegraphics[width=.7\linewidth]{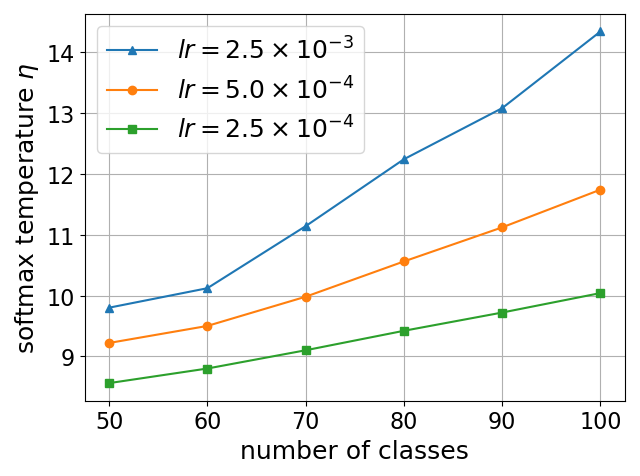}
	\caption{{\bf Comparison among different classifier learning rates} on learned softmax temperature $\eta$. Experiments conducted on ImageNet-100 B50 setting.(Best viewed in color)}
	\label{fig:LearningRate:Etavalue}
	\end{minipage}
	\hfill
	\begin{minipage}[c]{0.47\linewidth}
	\centering
	\includegraphics[width=.8\linewidth]{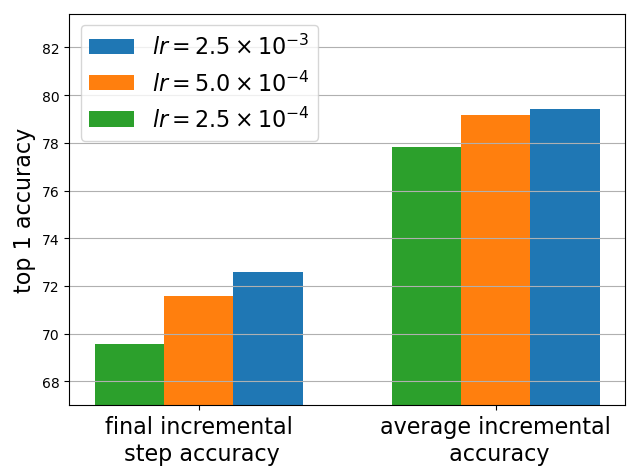}
	\caption{{\bf Comparison among different classifier learning rates} on average incremental accuracy, and accuracy of final incremental step. Experiments are conducted on ImageNet-100 B50 setting.(Best viewed in color)}
	\label{fig:LearningRate:Acc}
	\end{minipage}
\end{figure}

\vspace{2mm}
\noindent{\bf Underfitted classifier when using the same learning rate to ViT feature extractor:}
Another clear difference between ResNet and ViT is observed in the magnitude of learned softmax temperature $\eta$ in Equ.~\ref{equ:CosineSoftmax}: $\eta=34$ for ResNet vs. $\eta=10$ for ViT in the final model after the last incremental step. We hypothesize that (a) the magnitude of $\eta$ is correlated to the learning rate, and (b) the small learning rate $2.5\times 10^{-4}$ is good for ViT as feature extractor but too low for the classifier. To validate this, we conduct experiments by applying different learning rates ($2.5\times 10^{-4}$, $5\times 10^{-4}$, $2.5\times 10^{-3}$) for the classifier alone while keeping the learning rate of ViT feature extractor unchanged ($2.5\times 10^{-4}$). Experimental results validate our hypothesis. As shown in Fig.~\ref{fig:LearningRate:Etavalue}, $\eta$ is highly correlated to the learning rate. Fig.~\ref{fig:LearningRate:Acc} shows increasing learning rate from $2.5\times 10^{-4}$ to $2.5\times 10^{-3}$ for classifier boosts both the final and average incremental accuracy significantly.

\section{Our Method: ViTIL}
\label{sec:OurMethod}
In this section, we introduce our method ViTIL to address the aforementioned issues by using existing techniques, which significantly boosts the class incremental learning performance. We note that our contribution is \textit{not} in these techniques developed in previous works, but in nailing down the causes of ViT's degradation and effective treatments by leveraging existing techniques.
\subsection{Bag of Treatments in ViTIL}
\vspace{1mm}
\noindent \textbf{Convolutional stem for faster convergence:}
\begin{figure}[t]
	\centering
	\includegraphics[width=.65\linewidth]{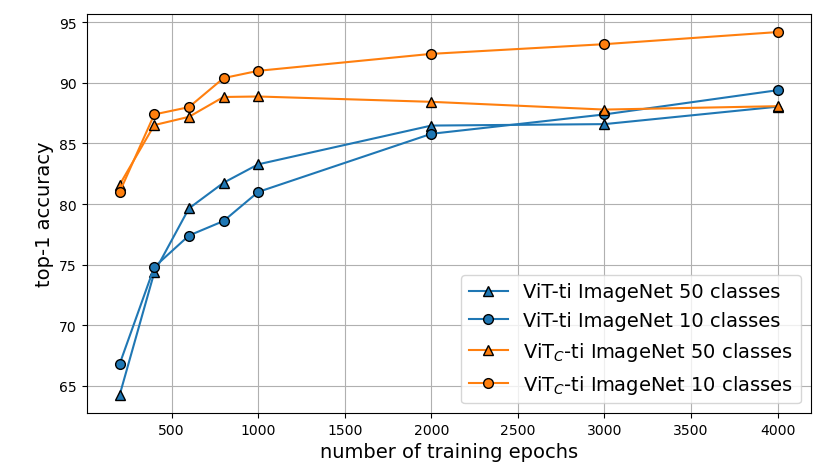}
	\caption{\textbf{Comparison between ViT and ViT$_C$} on accuracy over different number of training epochs. Experiments are conducted on the first 50 classes and 10 classes from ImageNet-100. (Best viewed in color)}
	\label{fig:DeiTConvstem}
\end{figure}
Inspired by the findings in~\cite{xiao2021early} that convolutional stem introduces quick converge, we follow it replace patchify operation and first transformer block with a small convolutional network with four convolution layers.
Each convolution layer has kernel size of 3, and stride size of 2. 
This change results in much faster convergence to saturated performance (see Fig.~\ref{fig:DeiTConvstem}).
For 50 classes from ImageNet-100, ViT with convolutional stem (denoted as ViT$_C$-ti) not only speeds up the convergence to around 800 epochs, but also achieves higher accuracy (+$11.68$\% and +$4.96$\%) at both 400 and 800 training epochs.
It outperforms ResNet-18 by $4.24$\%, providing a strong start for the following incremental steps.

\vspace{1mm}
\noindent \textbf{Bias correction via balanced finetuning:}
We correct the biases in classifier by finetuning it with balanced dataset, which is denoted as $\mathcal{X}^\prime_o \cup \mathcal{X}^\prime_n$, where $\mathcal{X}^\prime_o$ and $\mathcal{X}^\prime_n$ are the exemplar set from old and new classes respectively.
The idea of using balanced dataset to correct biases in the classifier was studied in~\cite{wu2019large}, where a linear layer with two parameters is learned. However, it does not apply to our case, since our classifier has normalized weights per class. Instead, we finetune all parameters in the classifier. As shown in Fig.~\ref{fig:ConfusionMatrixDeiTCBF}, this finetuning strategy effectively mitigates bias towards new classes for using ViT model.
This results in a two-stage incremental learning.
In the first stage, parameters in both feature extractor and classifier (denoted as $\Theta_t$ for incremental step $t$) are updated. The second stage deals with balance finetuning of classifier where only the parameters in classifier (denoted as $\Theta^{g}_t$) are finetuned, with feature extractor frozen.

\vspace{1mm}
\noindent \textbf{Large learning rate for classifier:}
It is straightforward to address the underfitting of classifier, by using a larger learning rate for it. In this paper, we use learning rate $2.5\times 10^{-3}$ on ImageNet dataset, which is ten times higher than the learning rate for the backbone feature extractor.

\subsection{Implementation Details}
\label{sec:ImpleDetails}
\noindent{\bf ViT architecture:} We follow the design of ViT models in~\cite{dosovitskiy2021image}, and adopt ViT-ti as the base model for ImageNet dataset.
Existing CIL works apply ResNet-18 as base model for CIL on ImageNet dataset.
ViT-ti has 5M parameters and 1.1G FLOPs, less than ResNet-18 with 12M parameters and 1.8G FLOPs.
CLS token of final transformer block is treated as the feature output of backbone ViT model, and fed into classifier head, which is cosine linear layer in our case.
For CIL on CIFAR-100 dataset, existing works apply a modified 32-layer ResNet with 0.47M parameters, 69.8M FLOPs.
For a fair comparison, we modify our ViT model with embedding dimension of 120, depth of 4, mlp ratio of 1, and patch size of 4.
This lightweight model contains 0.47M parameters, and 41.0M FLOPs, with input image resolution 32$\times$32 same as existing works.

\noindent{\bf Convolutional stem architecture:} 
Follow~\cite{xiao2021early}, we replace the patchify stem and the first transformer block in ViT-ti model with a small convolutional network.
It consists of 4 convolution layers with kernel size $3$ and stride size $2$.
The number of output channels is [24, 48, 96, 192].
Each convolution layer is followed by a batch norm layer and a ReLU layer.
As pointed out in~\cite{xiao2021early}, convolutional stem has negligible impact on the model's FLOPs.

\noindent{\bf Loss:}
Our loss function, $L=\sum_{x\in\mathcal{X}^\prime_o\cup \mathcal{X}_n}(L_{ce}(x)+\lambda L_{dis}(x))$, includes a standard cross entropy loss $L_{ce}$, and distillation loss $L_{dist}$,
where $\lambda$ is a balancing factor.
Same as in~\cite{hou2019learning}, $L_{dis}$ penalizes the change of feature.
$L_{dis}(x) = 1-\left\langle \bar{f}^\ast(x),\bar{f}(x)\right\rangle$, where $f^\ast(x)$ represents feature vector extracted by old model.
In the finetuning stage, only cross entropy loss $L_{ce}$ is calculated on new class exemplar set $\mathcal{X}_n^\prime$ and old class exemplar set $\mathcal{X}_o^\prime$.
With feature extractor frozen, only cosine linear classifier parameters are updated.
$\lambda$ is set to 3.
Same as~\cite{hou2019learning}, we use adaptive $\lambda$.

\noindent{\bf Augmentation:} We adopt the same augmentation recipe in~\cite{xiao2021early} for ImageNet dataset, including Mixup, CutMix, soft label, and AutoAugmentation.
All images are re-scaled to 224$\times$224.
For CIFAR-100 dataset, we use same data augmentation with images re-scaled to 32$\times$32.

\noindent{\bf Optimizer:} Note that in all experiments, all ViT-based models are trained from scratch, without using any pre-training or extra dataset.
We adopt commonly used AdamW optimizer.
Batch size is set to 1024.
Weight decay is set to 0.24.
Learning rate is set to $2.5\times 10^{-4}$ for feature extractor, and $2.5\times 10^{-3}$ for classifier.
Learning rate is scaled based on batch size as $lr\times$BatchSize/512.
Cosine learning rate scheduler is applied with 5 warmup epochs.
For the initial step $t=1$, the number of training epochs is set to $800$.
For the following incremental steps $t>1$, number of training epochs is set to $50$ for B50/B500 settings, and $200$ for B0 setting.
This is because when the incremental learning starts with half of total classes, learned feature representation is more robust and can better generalize to the other half classes.
For CIFAR-100 dataset, since it contains less training data, number of training epochs is set to $2000$ for initial step $t=1$, and $500$ for following incremental steps $t>1$.
Batch size is set to 512.
Learning rate is set to $2.5\times 10^{-3}$ for feature extractor, and $5\times 10^{-3}$ for classifier.

%% file: experiment.tex
\section{Expriments}
\label{sec:Exp}

In this section, we evaluate the proposed method with public benchmarks and commonly used protocols.
Ablation study is provided to analyze the impact of different components of the proposed method.

\subsection{Settings}
\label{sec:Exp:Setting}

\noindent{\bf Datasets:}
Our experiments are conducted on three widely used CIL evaluation datasets, ImageNet-1000~\cite{deng2009imagenet}, ImageNet-100 and CIFAR-100~\cite{krizhevsky2009learning}.
ImageNet-1000 is a large-scale dataset with 1000 classes.
ImageNet-100 is a subset of ImageNet-1000 with 100 classes.
Same as in~\cite{hou2019learning}, we use random seed 1993 to shuffle 1000 classes and choose the first 100 classes as ImageNet-100.
CIFAR-100 is a dataset with small image resolution $32\times32$.

\vspace{1mm}
\noindent{\bf Protocols:}
CIL protocols in existing works are different on three main factors: number of classes in the initial step, number of new classes each incremental step adds, and the number of exemplars.
For simplicity, we use abbreviated notation of CIL protocol as in Fig.~\ref{fig:Teaser} (b).
Existing works mainly adopt three CIL protocols.
For example, (1) B500 R20, (2) B0 R20, (3) B0 T20K on ImageNet-1000 dataset.
B500 denotes initial step contains half of total 1000 classes. 
B0 denotes model starts from scratch, and each step adds same number of new classes.  
R20 denotes each old class keeps same number of 20 exemplars. 
T20K denotes each incremental step keeps total 20000 exemplars.
These protocols are evaluated with different numbers of new classes each incremental step adds, \eg, C100 denotes each incremental step adds 100 new classes.
For ImageNet-100 and CIFAR-100, B500 is changed to B50, and T20K is changed to T2K.
We conduct experiments on C2, C5 and C10 settings for CIFAR-100, C10 and C5 settings for ImageNet-100, C100 setting for ImageNet-1000.

\vspace{1mm}
\noindent{\bf Baselines:}
In our experiments, we compare with 5 baselines, including iCaRL~\cite{rebuffi2017icarl}, LUCIR~\cite{hou2019learning}, BiC~\cite{wu2019large}, PODNet~\cite{douillard2020podnet}, and DDE~\cite{hu2021distilling}.
Note that these baselines may only provide results on part of the three protocols in their original papers.
Here, we report their reported results for the protocols they have evaluated.
For the protocols they do not evaluate, we use either their official code release, or our own implementation.
BiC~\cite{wu2019large} is evaluated with our PyTorch implementation.
Note that in the released code of PODNet~\cite{douillard2020podnet} for ImageNet, stride size of first convolution layer in ResNet-18 is changed from 2 to 1, which results in larger spatial size of every convolution feature map and much larger FLOPs (6.9G FLOPs vs 1.8G FLOPs).
To investigate this issue, we evaluated it with two settings.
In the first one, we adopt official ResNet-18 setting for feature extractor.
In the second one, we use the modified ResNet same as in~\cite{douillard2020podnet}.
Our method is also evaluated with these two settings. 

\begin{table*}[t]
	\centering
	\small
	\caption{{\bf Average incremental accuracy ($\%$)} on ImageNet. All baselines apply regular ResNet-18 with 1.8G FLOPs. Our method has 1.1G FLOPs. C10 denotes each incremental step adds 10 new classes. B50 denotes initial step contains 50 classes. R20 denotes each old class keeps 20 exemplars. T2K denotes each incremental step keeps total 2K exemplars. Results with $\ast$ are reported directly from original paper. BiC$^\dagger$ represents our implementation of BiC. PODNet(1.8G) represents changing feature extractor in PODNet to regular ResNet-18 with 1.8G FLOPs. Original PODNet in~\cite{douillard2020podnet} has 6.9G FLOPs.}
	\label{tab:2datasets}
	\setlength{\tabcolsep}{0.6mm}{
		\begin{tabular}{l|lll|lll|lll}
			\toprule
			\multirow{4}[0]{*}{\bf Methods} & \multicolumn{6}{c|}{\bf ImageNet-100} & \multicolumn{3}{c}{\bf ImageNet-1000} \\
			\cline{2-10}
			& \multicolumn{3}{c|}{C10} & \multicolumn{3}{c|}{C5} & \multicolumn{3}{c}{C100} \\
			\cline{2-10}
			& B50 & B0 & B0 & B50 & B0 & B0 & B500 & B0 & B0 \\
			& R20 & R20 & T2K & R20 & R20 & T2K & R20 & R20 & T20K \\
			\midrule
			\midrule
			iCaRL~\cite{rebuffi2017icarl} & 65.06 & 61.60 & 66.29 & 58.76 & 54.02 & 62.76 & 53.03 & 55.95 & 59.63 \\
			LUCIR~\cite{hou2019learning} & 70.47$^\ast$ & 55.33 & 62.02 & 68.09$^\ast$ & 45.72 & 56.80 & 64.34$^\ast$ & 57.14 & 60.65 \\
			BiC$^\dagger$~\cite{wu2019large} & 68.58 & 64.35 & 68.79 & 62.83 & 54.38 & 62.61 & 56.78 & 57.60 & 62.47 \\
			PODNet(1.8G)~\cite{douillard2020podnet} & 74.48 & 62.59 & 67.28 & 70.97 & 54.81 & 60.85 & 64.22 & 57.86 & 59.49 \\
			LUCIR+DDE~\cite{hu2021distilling} & 72.34$^\ast$ & 58.49 & 65.90 & 70.20$^\ast$ & 50.17 & 61.30 & 67.51$^\ast$ & 57.16 & 59.13 \\
			Ours & {\bf 79.43} & {\bf 69.68} & {\bf 73.09} & {\bf 76.92} & {\bf 61.57} & {\bf 67.17} & {\bf 69.20} & {\bf 65.13} & {\bf 68.75} \\
			\bottomrule
		\end{tabular}
	}
\end{table*}

\begin{table*}[t]
	\centering
	\small
	\caption{{\bf Average incremental accuracy ($\%$) on CIFAR-100}. C10 denotes each incremental step adds 10 new classes. B50 denotes initial step contains 50 classes. R20 denotes each old class keeps 20 exemplars. T2K denotes each incremental step keeps total 2K exemplars.}
	\label{tab:CIFAR}
	\setlength{\tabcolsep}{0.8mm}{
		\begin{tabular}{l|lll|lll|lll}
			\toprule
			\multirow{3}[0]{*}{\bf Methods} & \multicolumn{3}{c|}{\bf C10} & \multicolumn{3}{c|}{\bf C5} & \multicolumn{3}{c}{\bf C2}\\
			\cline{2-10}
			& B50 & B0 & B0 & B50 & B0 & B0 & B50 & B0 & B0 \\
			& R20 & R20 & T2K & R20 & R20 & T2K & R20 & R20 & T2K \\
			\midrule
			\midrule
			iCaRL~\cite{rebuffi2017icarl} & 57.25 & 60.02 & 64.74 & 53.47 & 55.72 & 63.94 & 48.02 & 47.67 & 60.91 \\
			LUCIR~\cite{hou2019learning} & 63.42$^\ast$ & 55.84 & 61.66 & 60.18$^\ast$ & 50.23 & 59.54 & 57.03 & 41.46 & 57.48 \\
			BiC$^\dagger$~\cite{wu2019large} & 59.24 & 60.68 & 65.58 & 53.64 & 56.60 & 63.66 & 47.77 & 45.80 & 60.10 \\
			PODNet~\cite{douillard2020podnet} & 66.02 & 56.61 & 56.55 & 63.70 & 49.74 & 49.43 & 60.39 & 40.98 & 40.84 \\
			LUCIR+DDE~\cite{hu2021distilling} & 65.27$^\ast$ & 57.09 & 60.40 & 62.36$^\ast$ & 50.50 & 55.45 & 54.07 & 36.37 & 53.08 \\
			Ours & {\bf 71.92} & {\bf 67.26} & {\bf 70.11} & {\bf 67.89} & {\bf 61.12} & {\bf 67.70} & {\bf 61.98} & {\bf 55.63} & {\bf 65.35} \\
			\bottomrule
		\end{tabular}
	}
\end{table*}

\subsection{Results}
\label{sec:Exp:Results}
\begin{figure*}[t]
	\centering
	\begin{subfigure}{0.325\linewidth}
		\includegraphics[width=\linewidth]{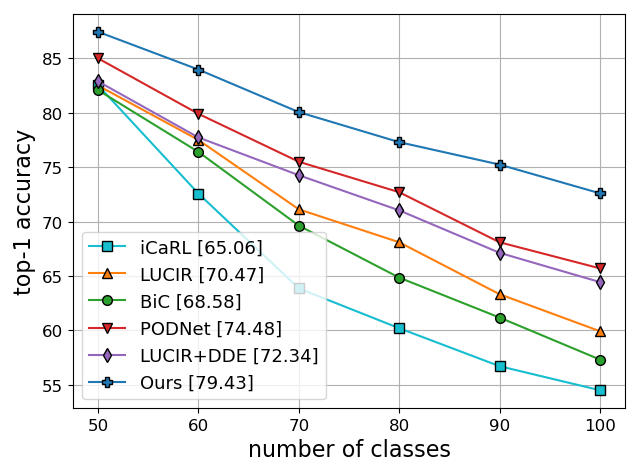}
		\caption{B50, C10, R20}
		\label{fig:fg_50_proto_20}
	\end{subfigure}
	\hfill
	\begin{subfigure}{0.325\linewidth}
		\includegraphics[width=\linewidth]{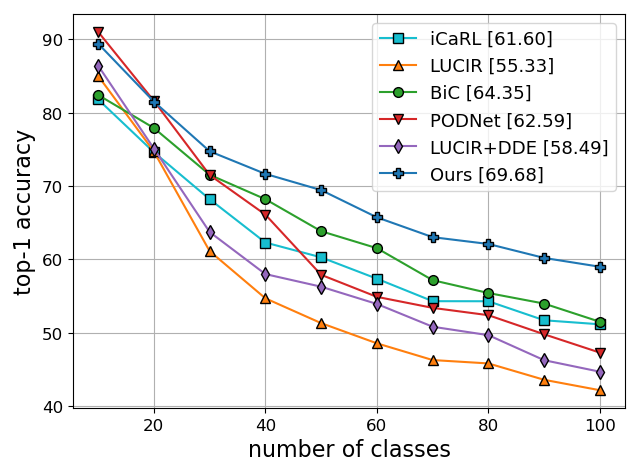}
		\caption{B0, C10, R20}
		\label{fig:fg_10_proto_20}
	\end{subfigure}
	\hfill
	\begin{subfigure}{0.325\linewidth}
		\includegraphics[width=\linewidth]{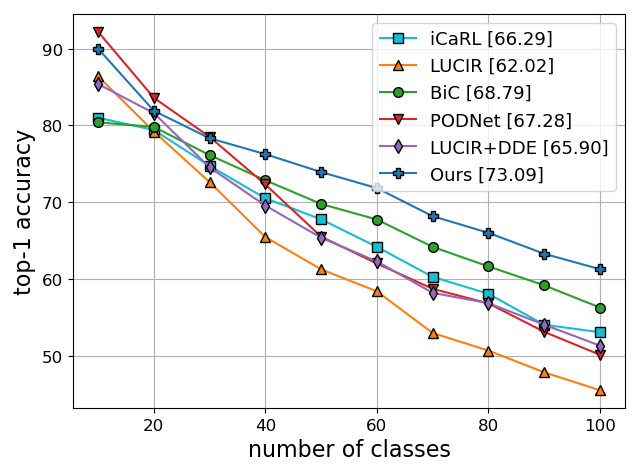}
		\caption{B0, C10, T2K}
		\label{fig:fg_10_proto_2000}
	\end{subfigure}
	\begin{subfigure}{0.325\linewidth}
		\includegraphics[width=\linewidth]{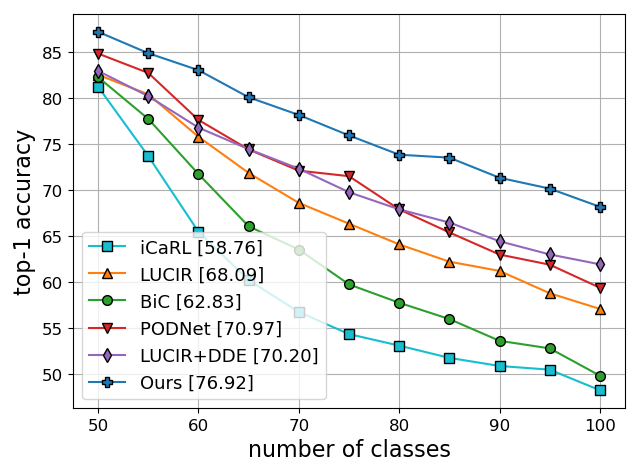}
		\caption{B50, C5, R20}
		\label{fig:fg_50_proto_20_inc5}
	\end{subfigure}
	\hfill
	\begin{subfigure}{0.325\linewidth}
		\includegraphics[width=\linewidth]{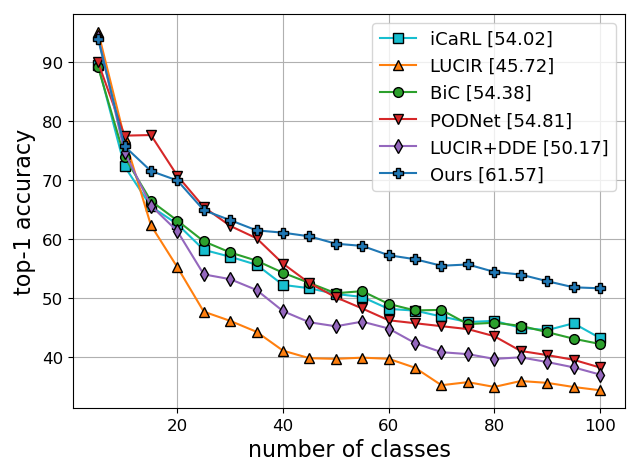}
		\caption{B0, C5, R20}
		\label{fig:fg_5_proto_20_inc5}
	\end{subfigure}
	\hfill
	\begin{subfigure}{0.325\linewidth}
		\includegraphics[width=\linewidth]{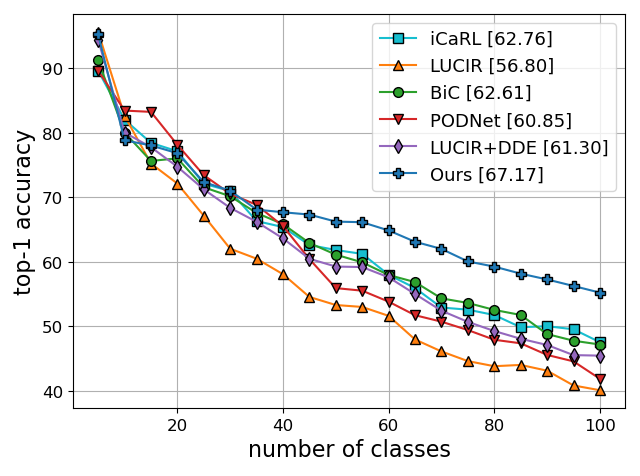}
		\caption{B0, C5, T2K}
		\label{fig:fg_5_proto_2000_inc5}
	\end{subfigure}
	\caption{Incremental accuracy of each incremental step in $\%$ on ImageNet-100. Number in $[\ ]$ denotes average incremental accuracy. (Best viewed in color)}
	\label{fig:IncAccCurve}
\end{figure*}

\noindent{\bf ImageNet-100:}
The quantitative results are shown in Tab.~\ref{tab:2datasets}.
The incremental accuracy of each incremental learning step is shown in Fig.~\ref{fig:IncAccCurve}.
Compared with CNN counter part, LUCIR~\cite{hou2019learning}, the proposed method outperforms by $8.96\%$, $14.35\%$, $11.07\%$, $8.83\%$, $15.85\%$, $10.37\%$ for 6 CIL settings on ImageNet-100, respectively.
Note that LUCIR, PODNet, BiC only perform well on partial CIL protocols.
In contrast, our method consistently outperforms existing methods on all CIL protocols.
Moreover, our accuracy degradation slope is flatter than CNN counterpart, LUCIR, indicating a better ability to address forgetting.
For example, with B50 C10 R20 in Fig.~\ref{fig:fg_50_proto_20}, accuracy improvement between ours and LUCIR is $4.94\%$ for the initial step, and $12.68\%$ for the final incremental step.

\begin{figure*}[t]
	\begin{subfigure}{0.325\linewidth}
		\includegraphics[width=\linewidth]{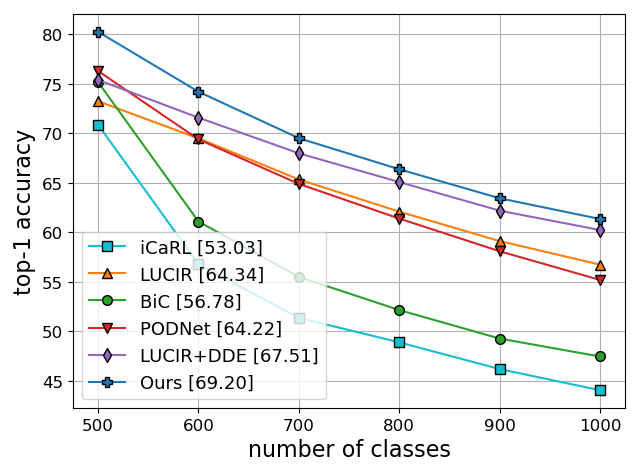}
		\caption{B500, C100, R20}
		\label{fig:fg_500_proto_20}
	\end{subfigure}
	\hfill
	\begin{subfigure}{0.325\linewidth}
		\includegraphics[width=\linewidth]{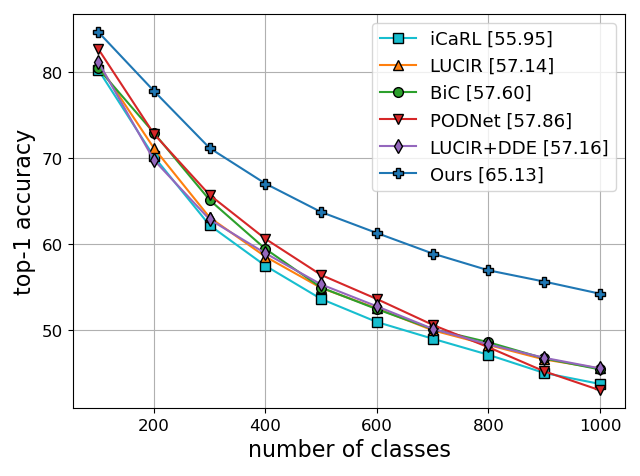}
		\caption{B0, C100, R20}
		\label{fig:fg_100_proto_20}
	\end{subfigure}
	\hfill
	\begin{subfigure}{0.325\linewidth}
		\includegraphics[width=\linewidth]{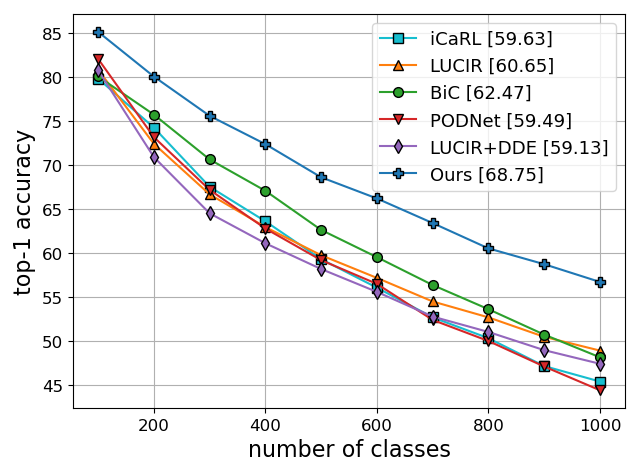}
		\caption{B0, C100, T20K}
		\label{fig:fg_100_proto_20000}
	\end{subfigure}
	\caption{Incremental accuracy of each incremental step in $\%$ on ImageNet-1000. Number in $[\ ]$ denotes average incremental accuracy. (Best viewed in color)}
	\label{fig:IncAccCurve1000}
\end{figure*}

\vspace{1mm}
\noindent{\bf ImageNet-1000:} 
As shown in Tab.~\ref{tab:2datasets}, the results on ImageNet-1000 is consistent with those on ImageNet-100.
With more classes, the catastrophic forgetting of ViT models is more severe compared with ImageNet-100.
The proposed method outperforms state-of-the-art methods by at least $1.69\%$ with B500 setting, by at least $7.27\%$ on B0 R20 setting, and by at least $6.28\%$ with B0 T20K setting.
Compared with CNN counterpart, baseline LUCIR, our improvements on three CIL protocols are $4.86\%$, $7.99\%$, and $8.10\%$, respectively.

\vspace{1mm}
\noindent{\bf CIFAR-100:}
The results are shown in Tab.~\ref{tab:CIFAR}.
Our method outperforms baselines with clear margin, using same input resolution 32$\times$32.
Moreover, our method has 0.47M parameters and 41.0M FLOPs, lower than 0.47M parameters 69.8M FLOPs of commonly adopted ResNet-32 model in existing works.
Note that our model is trained from scratch, without using any pre-training or extra dataset.

\begin{table}[t]
	\begin{minipage}[c]{0.47\linewidth}
	\centering
	\small
	\caption{\textbf{Comparing ViTIL with PODNet at higher FLOPs (6.9G)}. Average incremental accuracy ($\%$) on ImageNet-100 and ImageNet-1000. ``$\ast$" indicates the results reported in original paper~\cite{douillard2020podnet}.}
	\label{tab:largemodel}
	\setlength{\tabcolsep}{1.3mm}{
	\begin{tabular}{c|l|ll|ll}
		\toprule
		& \multicolumn{1}{c}{} & & & PODNet & Ours\\
		\cline{1-6}
		\multirow{6}{*}{\STAB{\rotatebox[origin=c]{90}{\footnotesize 100class}}} & \multirow{3}{*}{\STAB{\rotatebox[origin=c]{90}{C10}}} & B50 &  R20 & $75.54^\ast$ & {\bf 81.07} \\
		& & B0 & R20 & 67.01 & {\bf 70.41} \\
		& & B0 & T2K & 71.66 & {\bf 73.22} \\
		\cline{2-6}
		& \multirow{3}{*}{\STAB{\rotatebox[origin=c]{90}{C5}}} & B50 &  R20 & $74.33^\ast$ & {\bf 78.99} \\
		& & B0 & R20 & 58.77 & {\bf 62.85} \\
		& & B0 & T2K & 65.10 & {\bf 68.03} \\
		\hline
		\multirow{3}{*}{\STAB{\rotatebox[origin=c]{90}{\footnotesize 1Kclass}}} &\multirow{3}{*}{\STAB{\rotatebox[origin=c]{90}{C100}}} & B500 &  R20 & $66.95^\ast$ & {\bf 72.36} \\
		& & B0 & R20 & 61.60 & {\bf 67.40} \\
		& & B0 & T20K & 62.68 & {\bf 70.69} \\
		\bottomrule
	\end{tabular}
	}
	\end{minipage}
	\hfill
	\begin{minipage}[c]{0.47\linewidth}
	\centering
	\small
	\caption{Ablative results on ImageNet-100 dataset C10 setting. Average incremental accuracy of different variants. CNN denotes CNN stem. Bias denotes bias correction. Large lr denotes large classifier learning rate.}
	\label{tab:ablation}
	\setlength{\tabcolsep}{0.6mm}{
		\begin{tabular}{ccc|l|l|l}
			\toprule
			\multirow{2}[0]{*}{CNN} & \multirow{2}[0]{*}{Bias} & \multirow{2}[0]{*}{Large lr} & B50 & B0 & B0 \\
			& & & R20 & R20 & T2K \\
			\midrule
			\midrule
			& & & 67.11 & 45.17 & 50.02 \\
			& $\checkmark$ & & 70.07 & 54.67 & 57.19 \\
			& $\checkmark$ & $\checkmark$ & 74.13 & 59.45 & 61.70 \\
			\hline
			$\checkmark$ & & & 71.52 & 56.60 & 63.06 \\
			$\checkmark$ & $\checkmark$ & & 77.83 & 67.88 & 70.09 \\
			$\checkmark$ & $\checkmark$ & $\checkmark$ & {\bf 79.43} & {\bf 69.68} & {\bf 73.09} \\
			\bottomrule
		\end{tabular}
	}
	\end{minipage}
\end{table}

\vspace{1mm}
\noindent{\bf Model with larger FLOPs:}
Here we change the first convolution layer of our method to stride size 1, and obtain a model with 6.9G FLOPs.
In the released code of PODNet~\cite{douillard2020podnet}, the first convolution layer of ResNet-18 is modified with kernel size of 3 and stride size of 1.
With this modification, PODNet has 6.9G FLOPs.
For a fair comparison, we compare with our model with 6.9G FLOPs.
The results are shown in Tab.~\ref{tab:largemodel}.
Our method still outperforms PODNet consistently on ImageNet-100, ImageNet-1000 with all CIL protocols.

\begin{table}[t]
	\centering
	\small
	\caption{Average incremental forgetting ($\%$) on ImageNet-100. Lower is better.}
	\label{tab:forgetting}
	\setlength{\tabcolsep}{1.5mm}{
		\begin{tabular}{l|lll|lll}
			\toprule
			\multirow{3}[0]{*}{\bf Methods} & \multicolumn{3}{c|}{\bf C10} & \multicolumn{3}{c}{\bf C5} \\
			\cline{2-7}
			& B50 & B0 & B0 & B50 & B0 & B0 \\ 
			& R20 & R20 & T2K & R20 & R20 & T2K \\
			\midrule
			\midrule
			iCaRL~\cite{rebuffi2017icarl} & 21.88 & 28.95 & 22.60 & 24.64 & 34.80 & 26.11 \\
			LUCIR~\cite{hou2019learning} & 13.92 & 38.62 & 28.36 & 15.53 & 46.34 & 32.03 \\
			BiC$^\dagger$~\cite{wu2019large} & 10.50 & 18.92 & {\bf 12.45} & 10.79 & 24.90 & 16.08 \\
			PODNet(1.8G FLOPs)~\cite{douillard2020podnet} & 7.60 & 19.09 & 16.25 & 21.22 & 32.31 & 30.16 \\
			LUCIR+DDE~\cite{hu2021distilling} & 5.44 & 32.11 & 19.65 & 5.03 & 35.94 & 19.56 \\
			Ours & {\bf 2.37} & {\bf 18.01} & 13.92 & {\bf 2.83} & {\bf 17.34} & {\bf 12.25} \\
			\bottomrule
		\end{tabular}
	}
\end{table}

\subsection{Ablation Study}
\label{sec:Exp:Ablation}
In this section, we investigate the effectiveness of each component in the proposed method.
All variants apply same training hyper-parameters.
The results on ImageNet-100 are summarized in Tab.~\ref{tab:ablation}.
Moreover, we investigate the methods' ability to mitigate forgetting with evaluation metric of average incremental forgetting, same as in~\cite{hu2021distilling}.  

\vspace{1mm}
\noindent{\bf Convolutional stem:} As shown in Tab.~\ref{tab:ablation}, without convolutional stem, the naive ViT model's performance is significantly lower than its ResNet counterpart, LUCIR.
Moreover, with the same setting of classifier learning rate and bias correction, the models without CNN stem perform significantly worse than the models with CNN stem.
This is mainly due to the inferior optimizability of ViT models, which requires much longer training schedule.

\vspace{1mm}
\noindent{\bf Bias correction:} 
As shown in Tab.~\ref{tab:ablation}, without bias correction, the average incremental accuracy is much lower, especially for the protocol with more incremental steps with B0 protocols.
This is because bias towards new classes is more severe when there are more incremental steps.

\vspace{1mm}
\noindent{\bf Large classifier learning rate:} 
As shown in Tab.~\ref{tab:ablation}, when the model has bias correction, larger classifier learning rate further improves average incremental accuracy, which is caused by underfitted classifier parameters.
Moreover, it also improves ViT models without CNN stem.
In particular, without CNN stem, the initial step accuracy of ViT-ti is slightly below LUCIR on B50 C10 R20 setting.
However, with bias correction and large classifier lr, it still outperforms LUCIR on average incremental accuracy by $3.66\%$.

\vspace{1mm}
\noindent{\bf Average incremental forgetting:}
As shown in Tab.~\ref{tab:forgetting}, forgetting of our method is significantly lower than CNN counterpart, LUCIR.
Please note that lower forgetting does NOT necessarily lead to higher average accuracy, when initial accuracy is the same.
For example, in Tab.~\ref{tab:forgetting} B50 R20, forgetting of BiC is lower than LUCIR.
However, as in Fig.~\ref{fig:fg_50_proto_20}, average incremental accuracy of BiC is lower than LUCIR, and their initial step accuracy is very close.
This is because average forgetting only measures model stability, while CIL pursues a trade-off between stability and plasticity.
A trivial solution, with model frozen after initial step, can achieve 0 forgetting, while not handling new classes.

\subsection{Comparison with Methods of Dynamic Structure}
\label{sec:Exp:CompDER}
In this section, we discuss the difference between our approach and methods with dynamic structure, and compare their performance and model parameter.
In our experiments, all baselines and our method use a fixed model structure in each incremental step.
In contrast, a branch of CIL methods apply dynamic model structure, where the model structure changes in each incremental step.
For example, in DER~\cite{yan2021dynamically}, individual feature extractor is trained for each step and then concatenated together to avoid forgetting issue.
Its model size continuously grows as more classes come in.
Dynamic model structure for anti-forgetting is not our main focus.
However, our method does not conflict with dynamic model structure, and has the potential to utilize it.

We compare with DER with the reported results in~\cite{yan2021dynamically}.
For ImageNet-1000, with setting B0 C100 T20K, DER obtains average accuracy of $66.73\%$ with 14.52M average parameters.
In contrast, our method achieves much higher accuracy $70.69\%$, with much lower parameter size 5.31M.
This demonstrates the advantage of our method on large-scale dataset.
For ImageNet-100, with settings B50 C5 R20, and B0 C10 T2K, DER achieves average accuracy of $77.73\%$ and $76.12\%$, with average parameter 8.87M and 7.67M, respectively.
Our method achieves average accuracy $78.99\%$ and $73.22\%$, while maintains constant parameter size 5.31M.
More detailed comparison is provided in appendix.

%% file: conclusion.tex
\section{Conclusion}
\label{sec:Conclusion}
In this paper, we investigate applying ViT models to class incremental learning scenario.
We find that naively replacing CNN feature extractor of current CIL method with ViT model results in severe performance degradation.
We nail down the causes of this performance degradation, and further address these issues with simple and effective existing techniques.
Our proposed method consistently outperforms state-of-the-art class incremental learning methods on two benchmarks across different CIL settings, by a clear margin.
Ablation study also demonstrates the effectiveness of our method.
The proposed method provides a strong baseline for future class incremental learning research.
One drawback of current method is the training schedule still longer than standard ResNet training recipe, which is currently an unsolved issue of ViT models in the community.
More works will be done to address it in the future.

%% file: appendix.tex
\section{Additional Experimental Results}
In this section, we show additional experimental results on CIFAR-100.
In addition, we compare with methods using dynamic model structure, and CNN methods with more data augmentation.
\subsection{Results on CIFAR-100}
\begin{figure*}[t]
	\centering
	\includegraphics[width=\linewidth]{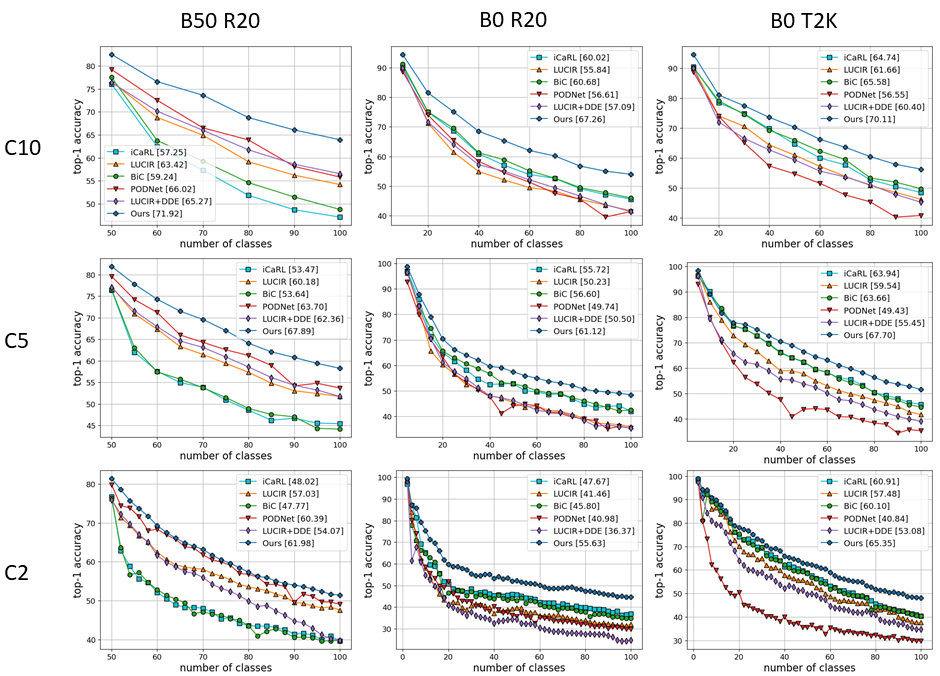}
	\caption{{\bf Incremental accuracy of each incremental step in $\%$ on CIFAR-100}. Number in $[\ ]$ denotes average incremental accuracy. The first row is C10 setting, where each incremental step adds 10 new classes. The second row and third row are C5 and C2 setting, respectively. The first column is B50 R20 setting, where initial step contains 50 classes, each old class keeps 20 exemplars. The second column and third column are B0 R20, and B0 T2K setting, respectively. (Best viewed in color)}
	\label{fig:IncAccCurveCIFAR100}
\end{figure*}

In this section, we provide experimental results on CIFAR-100.
Fig.~\ref{fig:IncAccCurveCIFAR100} shows the incremental accuracy of each incremental step with 9 CIL settings.
Our method outperforms all baselines by a clear margin on all CIL protocols.
Moreover, between our method and baselines, the accuracy improvement of each incremental step increases as more classes come in.
For example, for B0 C10 T2K setting, our improvement over LUCIR on initial step is 4.50\% (94.50\% vs 90.00\%), while the improvement on final incremental step is 10.04\% (56.28\% vs 46.24\%), and improvement of average incremental accuracy is 8.45\% (70.11\% vs 61.66\%).
This demonstrates that our performance boost of average incremental accuracy is larger than the boost of initial step.

\subsection{Comparison with Methods of Dynamic Structure}
In this section, we compare our method VITIL with methods using dynamic model structure.
A recent state-of-the-art method, DER~\cite{yan2021dynamically}, learns a separate feature extractor for new classes of each incremental step, and concatenate them together for classification.
As a result, its model structure changes in each incremental step, and its model size continuously grows as more classes come in.
In contrast, our method is based on a fixed model structure, which is simpler than DER.
Moreover, the model size is fixed in our method, while DER model size grows as more classes come in.
In addition, since DER applies model pruning, the model has irregular structure and random memory access.
To the contrary, our method is based on regular model structure.

We compare with the performance of DER reported in~\cite{yan2021dynamically}.
Tab.~\ref{tab:CompDER} shows comparison with DER on both average incremental accuracy and number of model parameters on ImageNet-100 and ImageNet-1000.
Tab.~\ref{tab:CompDERCIFAR} shows comparison on CIFAR-100.
Out of total 7 CIL setups, our method outperforms DER for 5 setups.
On the most challenging dataset, ImageNet-1000, our method achieves 3.96\% (70.69\% vs 66.73\%) improvement on average incremental accuracy with setting B0 C100 T20K.
Moreover, our model parameter size is 63.4\% smaller than DER (5.31M vs 14.52M).
This demonstrates the advantage of our method on large-scale dataset.
For ImageNet-100, with B50 C5 R20 setting, our performance improvement is 1.26\% (78.99\% vs 77.73\%), with model size 40.1\% smaller than DER (5.31M vs 8.87M).
For B0 C10 T2K setting, our model size is 30.1\% smaller.
For the 4 CIL protocols on CIFAR-100, our method outperforms DER on 3 protocols with smaller model size.

\begin{table*}[t]
	\centering
	\small
	\caption{{\bf Comparison between VITIL and DER on ImageNet-1000 and ImageNet-100} over average incremental accuracy ($\%$) and number of model parameters (M). Acc denotes average incremental accuracy. \#Paras denotes number of model parameters. C5 denotes each incremental step adds 5 new classes. B50 denotes initial step contains 50 classes. R20 denotes each old class keeps 20 exemplars. T2K denotes each incremental step keeps total 2K exemplars.}
	\label{tab:CompDER}
	\setlength{\tabcolsep}{1.5mm}{
		\begin{tabular}{l|l|l|l|l|l|l}
			\toprule
			\multirow{5}{*}{\bf Methods} & \multicolumn{2}{c|}{\bf ImageNet-1K} & \multicolumn{4}{c}{\bf ImageNet-100} \\
			\cline{2-7}
			& \multicolumn{2}{c|}{C100} & \multicolumn{2}{c|}{C5} & \multicolumn{2}{c}{C10} \\
			\cline{2-7}
			& \multicolumn{2}{c|}{B0} & \multicolumn{2}{c|}{B50} & \multicolumn{2}{c}{B0} \\
			& \multicolumn{2}{c|}{T20K} & \multicolumn{2}{c|}{R20} & \multicolumn{2}{c}{T2K} \\
			\cline{2-7}
			& Acc & \#Paras & Acc & \#Paras & Acc & \#Paras \\
			\midrule
			\midrule
			DER~\cite{yan2021dynamically} & 66.73 & 14.52 & 77.73 & 8.87 & {\bf 76.12} & 7.67 \\
			Ours & {\bf 70.69} & {\bf 5.31} & {\bf 78.99} & {\bf 5.31} & 73.22 & {\bf 5.31} \\
			\bottomrule
		\end{tabular}
	}
\end{table*}

\begin{table*}[t]
	\centering
	\small
	\caption{{\bf Comparison between VITIL and DER on CIFAR-100} over average incremental accuracy ($\%$) and number of model parameters (M). Acc denotes average incremental accuracy. \#Paras denotes number of model parameters. C10 denotes each incremental step adds 10 new classes. B50 denotes initial step contains 50 classes. R20 denotes each old class keeps 20 exemplars. T2K denotes each incremental step keeps total 2K exemplars.}
	\label{tab:CompDERCIFAR}
	\setlength{\tabcolsep}{1.3mm}{
		\begin{tabular}{l|l|l|l|l|l|l|l|l}
			\toprule
			\multirow{4}{*}{\bf Methods} & \multicolumn{4}{c|}{{\bf C10}} & \multicolumn{4}{c}{{\bf C5}} \\
			\cline{2-9}
			& \multicolumn{2}{c|}{B50} & \multicolumn{2}{c|}{B0} & \multicolumn{2}{c|}{B50} & \multicolumn{2}{c}{B0} \\
			& \multicolumn{2}{c|}{R20} & \multicolumn{2}{c|}{T2K} & \multicolumn{2}{c|}{R20} & \multicolumn{2}{c}{T2K} \\
			\cline{2-9}
			& Acc & \#Paras & Acc & \#Paras & Acc & \#Paras & Acc & \#Paras \\
			\midrule
			\midrule
			DER~\cite{yan2021dynamically} & 67.60 & 0.59 & 69.41 & 0.52 & 66.36 & 0.61 & {\bf 68.82} & {\bf 0.45} \\
			Ours & {\bf 71.92} & {\bf 0.47} & {\bf 70.11} & {\bf 0.47} & {\bf 67.89} & {\bf 0.47} & 67.70 & 0.47 \\
			\bottomrule
		\end{tabular}
	}
\end{table*}

\subsection{CNN with More Data Augmentation}
In this section, we investigate if CNN-based method with same data augmentation in ViT model can achieve similar incremental learning performance.
We add same data augmentation of ViT to LUCIR with ResNet-18, including Mixup, CutMix, soft label, and AutoAugmentation.
Similarly, margin ranking loss does not apply anymore due to soft label, and is replaced by class balance finetuning.
Results on ImageNet-100 and ImageNet-1000 are shown in Tab.~\ref{tab:LUCIRwDataAug}.
It does not show improvements over original LUCIR.

\begin{table*}[t]
	\centering
	\small
	\caption{{\bf Comparison between LUCIR w/ and w/o data augmentation in ViT models}. Average incremental accuracy ($\%$) on ImageNet-100 ImageNet-1000. LUCIR+Aug denotes adding same data augmentation from ViT to LUCIR with ResNet-18. Results with $\ast$ are reported directly from original paper.}
	\label{tab:LUCIRwDataAug}
	\setlength{\tabcolsep}{1.3mm}{
		\begin{tabular}{l|lll|lll|lll}
			\toprule
			\multirow{4}[0]{*}{\bf Methods} & \multicolumn{6}{c|}{\bf ImageNet-100} & \multicolumn{3}{c}{\bf ImageNet-1000} \\
			\cline{2-10}
			& \multicolumn{3}{c|}{C10} & \multicolumn{3}{c|}{C5} & \multicolumn{3}{c}{C100} \\
			\cline{2-10}
			& B50 & B0 & B0 & B50 & B0 & B0 & B500 & B0 & B0 \\
			& R20 & R20 & T2K & R20 & R20 & T2K & R20 & R20 & T20K \\
			\midrule
			\midrule
			LUCIR~\cite{hou2019learning} & 70.47$^\ast$ & 55.33 & 62.02 & 68.09$^\ast$ & 45.72 & 56.80 & 64.34$^\ast$ & 57.14 & 60.65 \\
			LUCIR + Aug & 69.26 & 50.65 & 55.30 & 66.25 & 43.73 & 53.80 & 62.91 & 52.55 & 53.69 \\
			\bottomrule
		\end{tabular}
	}
\end{table*}